\documentclass[10pt,twocolumn,letterpaper]{article}

\usepackage[pagenumbers]{cvpr} % To force page numbers, e.g. for an arXiv version

\definecolor{cvprblue}{rgb}{0.21,0.49,0.74}
\usepackage[pagebackref,breaklinks,colorlinks,citecolor=cvprblue]{hyperref}

\def\paperID{23} % *** Enter the Paper ID here
\def\confName{3DV\xspace}
\def\confYear{2025\xspace}

\title{LSSInst: Improving Geometric Modeling in LSS-Based BEV Perception with Instance Representation}

\author{
Weijie Ma\textsuperscript{1~\ddag}
\quad Jingwei Jiang\textsuperscript{1~\ddag}
\quad Yang Yang\textsuperscript{1} 
\quad Zehui Chen\textsuperscript{2}
\quad Hao Chen\textsuperscript{1} \thanks{~is the corresponding author and \ddag~denotes equal contribution.}\\
\textsuperscript{1}Zhejiang University \quad 
\textsuperscript{2}University of Science and Technology of China
}

\begin{document}
\maketitle
\begin{abstract}
With the attention gained by camera-only 3D object detection in autonomous driving, methods based on Bird-Eye-View (BEV) representation especially derived from the forward view transformation paradigm, i.e., lift-splat-shoot (LSS), have recently seen significant progress. The BEV representation formulated by the frustum based on depth distribution prediction is ideal for learning the road structure and scene layout from multi-view images. However, to retain computational efficiency, the compressed BEV representation such as in resolution and axis is inevitably weak in retaining the individual geometric details, undermining the methodological generality and applicability. With this in mind, to compensate for the missing details and utilize multi-view geometry constraints, we propose LSSInst, a two-stage object detector incorporating BEV and instance representations in tandem. The proposed detector exploits fine-grained pixel-level features that can be flexibly integrated into existing LSS-based BEV networks. Having said that, due to the inherent gap between two representation spaces, we design the instance adaptor for the BEV-to-instance semantic coherence rather than pass the proposal naively. Extensive experiments demonstrated that our proposed framework is of excellent generalization ability and performance, which boosts the performances of modern LSS-based BEV perception methods without bells and whistles and outperforms current LSS-based state-of-the-art works on the large-scale nuScenes benchmark. Code is available at \href{https://github.com/WeijieMax/LSSInst}{https://github.com/WeijieMax/LSSInst}.
\end{abstract}    
\section{Introduction}
\label{sec:intro}

As a crucial component in 3D perception, 3D object detection can be applied in various fields, such as autonomous driving and robotics. Although LiDAR-based 3D detection methods  \cite{ vora2020pointpainting, lang2019pointpillars, zhou2018voxelnet, SECOND18} are verified as having remarkable performance, research in camera-based methods 
\cite{LSS20, FCOS3D, DETR3D21, huang2021bevdet, huang2022bevdet4d, BEVFormer22} 
has received increasing attention in recent years. The reasons can be attributed not only to the lower deployment cost but also to the advantages offered by long-range distance and the identification of visual road elements \cite{BEVFormer22, SOLOFusion23}. However, unlike LiDAR sensors that provide direct and accurate depth information, detecting objects solely based on camera sensor images poses a significant challenge. Thus, how to utilize multi-view images to build up effective representations has become a key issue.

\begin{figure}[t]
  \centering
  \includegraphics[width=0.95\linewidth]{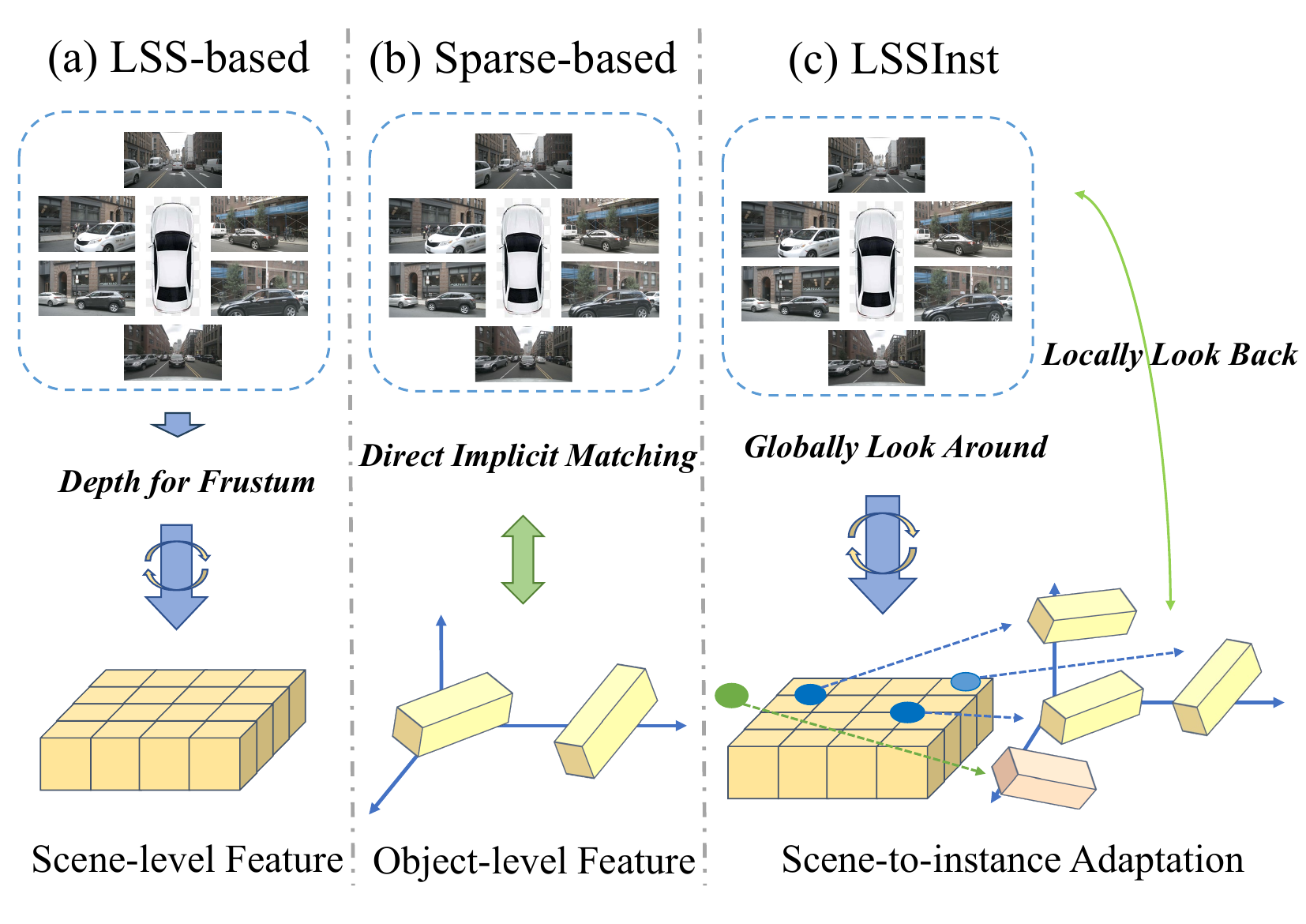}
  \caption{The conceptual comparison of LSSInst with previous camera-based fashions.}
  \label{concept}
\end{figure}

Recently, significant progress has been achieved in methods that utilize the bird's-eye view (BEV) whose view transformation can be mainly categorized as forward type \cite{BEVDepth23, BEVStereo23, aedet23, SOLOFusion23} based on lift-splat-shoot (LSS) \cite{LSS20} and backward type \cite{BEVFormer22, BEVFormerV2} based on the learnable BEV query. Due to its purely implicit aggregation by uninterpretable but forcibly dense queries, the backward type shows lower performance and expansibility \cite{videobev2023}, enabling the LSS-based forward type to become mainstream in modern BEV paradigms for camera-only 3D detection at present. Based on the LSS hypothesis and the fact that most objects in the scene are close to the ground, LSS-based BEV provides a perspective with minimal parallax ambiguity and information loss in observing the objects as a whole. Illustrated by Fig. \ref{concept} (I), these methods lift images into frustums based on depth distribution prediction and splat them into a BEV space, gathering information from multiple 2D views for a global representation of the scene. This representation is in the form of a planar view with compressed height ($z$-axis) and reduced resolution to ensure computational efficiency. The BEV feature benefits from its holistic representation and dense feature space, making it well-suited for capturing the scene's structure and data distribution. 
However, the geometrically-compressed nature of the BEV representation such as resolution and axis reduction, inherently limits its ability to provide precise 3D position descriptions of objects or fully exploit detailed features for object matching particularly in the 3D detection task which requires accurate prediction of 3D object bounding boxes.
Meanwhile, as another typical fashion and shown in Fig. \ref{concept} (II), sparse-based methods \cite{DETR3D21, PETR22, vedet} skip the BEV formulation and directly leverage object-level representations and exploit the 3D geometric prior to regress object bounding boxes from the 3D perspective. However, due to the initialization semantic dispersity \cite{li2022dndetr}, especially in more complex 3D perception, they fail to capture abundant object-aware features from the image at once in comparison with the BEV feature that fits adequate semantic information in the scene, resulting in lower overall performance than the contemporaneous BEV-based methods. 

\begin{table}[t]
\centering
\caption{The per-category AP comparison between two typical fashions with equivalent detection ability ($\Delta$mAP less than \textbf{0.5\%}) methods on the nuScenes $\mathtt{test}$ set.}
\label{class_differ}
\resizebox{\linewidth}{!}{
\begin{tabular}{c|c|c|c|c}
\toprule
 & \multicolumn{2}{c|}{\textbf{Group 1}} & \multicolumn{2}{c}{\textbf{Group 2}} \\
\midrule
 \multicolumn{1}{c|}{\textbf{Methods}} & \textbf{BEVDet \cite{huang2021bevdet}} & \textbf{Spatial-DETR \cite{Spatial-DETR}} & \textbf{CFT-BEV3D \cite{CFTBEV}} & \textbf{DETR3D \cite{DETR3D21}} \\
\midrule
\textbf{mAP} & 42.4 & 42.5 & 41.7 & 41.2 \\
\midrule
car & \textbf{0.643} & 0.610 & \textbf{0.628} & 0.603 \\
truck & \textbf{0.350} & 0.332 & \textbf{0.348} & 0.333 \\
bus & \textbf{0.358} & 0.330 & \textbf{0.347} & 0.290 \\
pedestrian & 0.411 & \textbf{0.462} & 0.416 & \textbf{0.455} \\
bicycle & 0.296 & \textbf{0.327} & 0.299 & \textbf{0.308} \\
traffic\_cone & 0.601 & \textbf{0.629} & 0.596 & \textbf{0.627} \\
barrier & \textbf{0.614} & 0.582 & \textbf{0.607} & 0.565 \\
\bottomrule
\end{tabular}
}
\end{table}

Following this, there are some interesting and corroborative findings in the per-class AP comparison between the two aforementioned fashions as shown in Tab. \ref{class_differ}. Notably, considering the practical variety such as data augmentation and training strategies, the difference between the overall mAP values of selected methods in the same group is strictly less than \textbf{0.5\%} which ensures both detection abilities are equivalent. We can observe that there is the same AP tendency among the classes. Specifically, BEV representation seems more attentive to regular objects (car, bus, truck, barrier) with distinct movements or common positions in the scene, with relative insensitivity to the objects (pedestrian, bicycle, traffic cone) with uncertain trajectories or dispersed locations, which further proves its characteristics of fitting data distribution and leaning to the scene-level focus. 
% is there a way to brighten the complementary synergy of both? Therefore, in order to
Inspired by this, to brighten the complementary synergy of both fashions and make up for the missing details in the representation formulation of current LSS-based BEV perception as well as utilize multi-view geometry constraints, we are motivated to propose \textbf{LSSInst}, incorporating the sparse instance-level representations based on the scene-level representations to look back for more detailed feature with geometric matching. As illustrated in Fig. \ref{concept} (III), based on the global scene-level pre-feature, instance-level features are pushed to look back at the image locally, focusing on more fine-grained pixel features and allowing for flexible geometric matching, which ultimately generates a final perception result that combines globally-semantic and locally-geometric information.

However, this collaboration also poses challenges, as the most straightforward solution of naively sharing the bounding box proposal is intuitively and experimentally failed \footnote{See Sec. \ref{abalation} and Tab. \ref{table:abl-Q} for more details}. As aforementioned, traditional sparse-based detection methods suffer from initialization semantic dispersity and inadequate semantic understanding of the scene, the above solution would sever the coherence with the dense representations. With this in mind, we propose the instance adaptor module to establish semantic coherence between the scene and instances and an instance branch for detection. The instance adaptor module generates multiple sparse queries and their corresponding 3D boxes through multi-level adaptive aggregation. The instance branch focuses on fine-grained sparse feature extraction and geometric matching using prepared inputs, such as box embeddings and spatiotemporal sampling and fusion. On the nuScenes dataset, our LSSInst method demonstrates strong generalization ability. Compared to other typical LSS-based methods, LSSInst achieves significant improvements in mAP.
Specifically, it outperforms BEVDet by 5.0\%, BEVDepth by 2.2\%, BEVStereo by 2.6\%, and 
surpasses the state-of-the-art LSS-based method SOLOFusion by 1.6\%.

Our main contributions can be concluded as follows: i) We proposed LSSInst, a two-stage framework that improves the geometric details in LSS-based BEV perception with instance representations; ii) We proposed the instance adaptor to maintain the BEV-to-instance semantic coherence and a newly-designed instance branch to look back and aggregate features spatiotemporally for improvement; iii) The proposed framework was verified with great generalization ability and surpassed the state-of-the-art LSS-based methods by extensive experimental results.

\section{Related Work}
\label{sec:related_work}

\subsection{LSS-based BEV Perception}
As BEV has proved to be an effective representation for multi-view 3D detection, LSS-based methods that benefit from the explicit formulation process and superior performance become the recent mainstream paradigm. LSS \cite{LSS20} is proposed for an end-to-end view transform architecture that lifts images into frustums by predicting depth distribution and splats them into a BEV representation. 
Then BEVDet \cite{huang2021bevdet} incorporates exclusive data augmentation techniques for the detection extension.
BEVDepth \cite{BEVDepth23} and BEVStereo \cite{BEVStereo23} improved the depth accuracy by introducing an extra monocular depth network supervised by corresponding LiDAR depth, and multi-view stereo matching between adjacent frames, while BEVDistill \cite{bevdistill} chose to the model-level distillation from LiDAR.
OA-BEV \cite{OA-BEV} and SA-BEV \cite{SA-BEV} enhanced the utilization of depth, which integrated the 3D voxel network based on the additional proposal from the 2D detection network and proposed a depth and semantic fusion module respectively for a more enhanced feature. 
Besides, several works started to perceive the shortage of the current view transformation assumption.
AeDet \cite{aedet23} introduced the positional compensation for existing coordinate projection while FB-BEV \cite{FB-BEV} integrated a novel forward-backward view transformation module that partially alleviates the projection issues.
SOLOFusion~\cite{SOLOFusion23} further unified long-term temporal information based on the short-term temporal optimization with Gaussian top-k sampling to boost performance.

Despite these methods making efforts to chase a flawless BEV representation from the LSS process, due to the avoidable depth error and the compressed nature of pooling operations, the yielded BEV representation is weak in retaining the individual geometric details, hence we differently focus on adapting the BEV representation into better geometrical modeling.

\subsection{Instance-level Representation Integration in Camera 3D Detection}
Integrating instance-level representation is ubiquitous in camera 3D detection to enhance perceptual ability. FQNet \cite{FQNet} is a three-stage framework for monocular detection that first locally searches for potential boxes and then follows a Fast-RCNN-like way \cite{girshick2015fast} to aggregate the massive object candidate globally for location prediction. \cite{li2022stereo} apply a similar way in stereo 3D detection. They first borrow DSGN \cite{DSGN} to locally search for possible proposals and then establish the Vernier network to globally form the confidence map based on the stereo pair. Similarly in multi-view 3D detection, for the sparse-based methods \cite{DETR3D21, PETR22} which solely depend on the query decoding, there tends to be a weak correlation and slow convergence between foreground tokens and queries due to the perspective inconsistency. Focal-PETR \cite{focal-petr} adopts extra 2D instance-level supervision to adaptively focus object queries on discriminative foreground regions. For BEV query-based methods, BEVFormer V2 \cite{BEVFormerV2} used the extra 3D perspective network such as \cite{DD3D, DETR3D21} to generate coarse instance features to serve as auxiliary proposals. Unlike they borrow instance-level features in a \textbf{bottom-up (i.e., local-to-global)} way, LSSInst uses a totally different \textbf{top-down (i.e., global-to-local)} way for improvement.

\subsection{Two-stage 3D Object Detector}
The two-stage design has been widely explored in the 3D detection domain and proved to be effective, whose multi-step workflow is favorable for more accurate prediction. 
For LiDAR 3D detection, drawing inspiration from 2D detection~\cite{girshick2015fast, ren2016faster, cai2018cascade}, a two-stage LiDAR detector typically generates Regions of Interest (RoIs) in the first stage and then refines these initial predictions in the second stage \cite{shi2019pointrcnn, yang2019std, qi2017pointnet++}. 
To address this issue of losing the ability to encode the geometric information of the proposals, point cloud pooling operations on the RoI \cite{shi2020part} or virtual points with boundary offsets \cite{li2021lidar} were introduced.
Instead of pooling from point features, Voxel R-CNN \cite{deng2021voxel} designs a voxel-RoI pooling module to pool directly from voxel and BEV feature space according to the RoI-grid. To increase speed, CenterPoint~\cite{centerpoint} simplifies the pooling module by sampling five keypoints from BEV features using bilinear interpolation while RSN~\cite{Sun_2021_CVPR} uses a foreground segmentation as a first stage to sparsify the point clouds, thereby enhancing the efficiency of the second stage sparse convolution. 
For camera 3D detection,
MonoDIS~\cite{DisentangleMono3D22} extracts features from 2D bounding boxes for subsequent 3D bounding box regression and introduces a disentangling transformation to supervise the detection loss separately for 2D and 3D.
SimMod~\cite{SimMod23} utilized a DETR3D head \cite{DETR3D21} to iteratively refine 2D-level object proposals output from a monocular network.
BEVFormer v2~\cite{BEVFormerV2} extends BEVFormer \cite{BEVFormer22} into the second stage by incorporating a first-stage 3D perspective detection network that directly supervises the image backbone network, leveraging both BEV and perspective information.
Among current two-stage methods, they seek more refinement for jointly aggregating coarse samples because their first stage primarily relies on perspective views. Instead, we focus on the subsequent refinement of proposals with holistic semantics derived together from the scene-level layout. 
% Moreover, compared to the above work including LiDAR detection, in the modeling of BEV and object query, we fully consider the object occlusion relationship in the scene that is unique to image data under a wide field of view. In the track of camera-only 3D detection, in spite of the inherent errors since image data cannot directly reveal the depth information, we improve this by the instance adapter and instance-level secondary optimization.
\section{Methodology}

\begin{figure*}[t]
  \centering
  \includegraphics[width=0.98\linewidth]{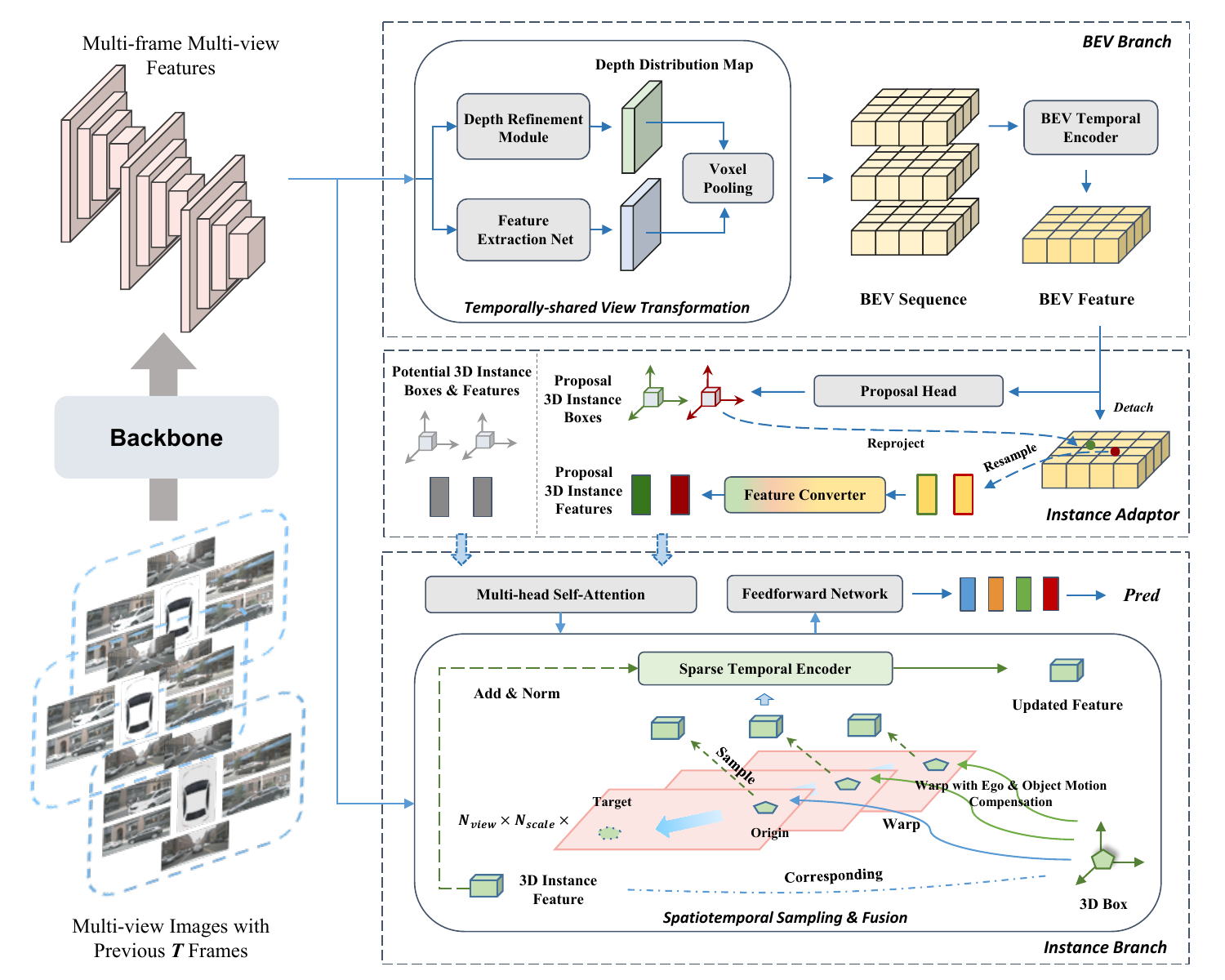}
  \caption{Overview of LSSInst. The multi-view images with previous $T$ frames are fed into the backbone network for the image features. BEV branch looks around the image feature to generate the BEV feature by view transformation and temporal encoding. Instance adapter aggregates the sparse object-aware feature from the BEV feature and prepares the multiplicate 3D query combination. Instance branch looks back at the image feature and perfects the sparse feature by spatiotemporal sampling and fusion. Lastly, the model makes the final prediction based on the updated output.}
  \label{overview}
\end{figure*}

Utilizing instance-level representations on the basis of the scene-level BEV to excavate more detailed features and geometric information is of practical significance for generalized 3D perception. In this work, we propose LSSInst, which looks back for the more geometry-aware and fine-grained target feature extraction to bridge the adaptation between scene-level and instance-level 3D representations. The overview of our framework is shown in Fig. \ref{overview}, and we organize the remaining part as follows. Firstly, Sec. \ref{BEVBranch} briefly introduces the BEV branch. Next, Sec. \ref{Bridge} introduces the instance adapter module, and the instance branch is given in Sec. \ref{SparseBranch}.

\subsection{BEV Branch: Looking around for scene-level representation}
\label{BEVBranch}
The multi-view sequential images with the previous $T$ frames are first input into the 2D image backbone network for feature extraction. Then the BEV branch receives the extracted image feature $F_{img} \in \mathbb{R}^{N_{v}\times N_{s}\times C\times H \times W}$ and functions as an around looker, translating $F_{img}$ from 2D camera views to BEV for preliminary scene-level representation $F_{\beta} \in \mathbb{R}^{C \times H_{\beta} \times W_{\beta}}$, where $N_v$, $N_s$ denote the camera-view and scale number respectively. This branch can be briefly divided into temporally-shared view transformation for BEV generation and BEV sequence fusion. The 2D-to-BEV view transformation is naturally based on LSS paradigm, which can be mainly concluded as the depth refinement module (DRM), feature extraction net, and voxel pooling. For the best version of the framework, we adopted the Gaussian-spaced top-k stereo \cite{SOLOFusion23} for a better depth distribution map before voxel pooling. After the shared view transformation, a sequence of BEV representations will be aligned into current time $t$ and fed to the BEV temporal encoder to form the final current BEV. Here the encoder is designed as a very lightweight residual network for dimension reduction only.

\subsection{Instance Adapter: Scene-to-instance adaptation}
\label{Bridge}

For the sake of preserving a coherent and solid semantic consistency between BEV and instance representations, we propose the instance adapter module to eliminate the gaps in the position description and space discrepancy. Since the BEV feature $F_{\beta}$ is a scene-level representation surrounding the ego car, there is redundancy and inflexibility in modeling instance-level features. To that end, the proposed adapter module first performs a reprojection of the proposal box coordinates $P_{o} \in \mathbb{R}^{N_{\beta} \times 3}$ obtained through the BEV proposal head, returning to the BEV-recognized position $P_{\beta} \in \mathbb{R}^{N_{\beta} \times 2}$ to resample the object-related features. Here $N_{\beta}$ denotes the number of BEV proposals. 
Given the BEV point-cloud range $R_{\beta}$ with the corresponding voxel size $S_v$ and up-sampling factor $\sigma$, we can formulate the 2D reprojected coordinate as follow:

\begin{equation}
\label{bev_reproject}
[P_{\beta}, z_{\beta}] = (P_{o} - R_{\beta})/(\sigma S_v)
\end{equation}
here $z_{\beta}$ denotes the z-axis homogeneous term and is actually a constant of 1. Moreover, due to the overfitting bias in BEV, the focused area may deviate from the actual object position. Inspired by deformable attention \cite{DeformableDETR20}, the adapter module incorporates the learnable offsets based on the original focused feature $F_{\beta}^{'} \in \mathbb{R}^{N_{\beta}\times C}$ for misalignment compensation by exploring more semantically-aware regions. Suppose $i\in \{1, 2, ..., N_{\beta}\}$ denotes an arbitrary element index of $F_{\beta}^{'}$ and its aggregated instance-wise feature $F_{\alpha}^i$ can be formulated by

\begin{equation}
\label{bev_deformable}
F_{\alpha}^i = \textbf{W}_{\alpha}\sum_{k = 1} ^K A_{k}^i \cdot \textbf{W}^{'}_{\alpha} \Phi \left(F_{\beta}, P_{\beta}^i + \Delta P_{\beta k}^i\right)
\end{equation}
where $\textbf{W}_{\alpha}, \textbf{W}^{'}_{\alpha} \in \mathbb{R}^{C \times C}$ are the weight matrix of linear projections and $k$ indexes the resampled keys with the total resampled key number $K$ ($K << H_{\beta}W_{\beta}$). $\Delta P_{\beta k}^i \in \mathbb{R}^2$ is the learnable offset and $A_{k}^i$ is the normalized attention weight of the $k^{th}$ sampling point with $\sum_{k = 1} ^K A_{k}^i = 1$, which are both linearly projected over $F_{\beta}^{'}$. $\Phi$ denotes the bilinear interpolation function.

Having said that aforementioned, there still remains the inherent space discrepancy between BEV encoding space and the 3D sparse space suitable for looking back at the image feature. Therefore, we first introduce an extremely shallow convolutional feature converter $f_{cnv}$ to reparameterize the aggregated features for the inter-space narration. Meanwhile, even with the extensive aggregation and enhancement based on BEV attentive features, a portion of irregular or separated objects cannot be detected due to the BEV overfitting of regular objects and relatively rough perception granularity. Therefore, we introduce extra learnable queries $F_{\gamma} \in \mathbb{R}^{N_{\gamma} \times C}$ and reference boxes independent of BEV proposals, named potential 3D instances and boxes, aimed at capturing the potential BEV-insensitive objects and learning a BEV-agnostic 3D spatial prior. Thus, we can get the multiplicate sparse feature $F_{\chi} \in \mathbb{R}^{N\times C}$ and here we let $N=N_{\beta}+N_{\gamma}$ for simplicity. The whole $F_{\chi} $ formulation can be derived by

\begin{equation}
\label{query_composition}
% F_{\chi} = \{f_{cnv}(\{F_{\alpha}^1, ..., F_{\alpha}^i, ..., F_{\alpha}^{N_{\beta}}\}), F_{\gamma}\}
F_{\chi} = \left \{f_{cnv} \left (\{F_{\alpha}^i\}^{N_{\beta}}_{i=1} \right ), F_{\gamma} \right \}
\end{equation}

\subsection{Instance Branch: Looking back for instance-level representation}
\label{SparseBranch}
 Given the sequential image features $\{F_{img}^t\}_{t=0}^{T_{\chi}}$ $(T_{\chi} \leq T)$ from the image backbone network and the sparse instance features $F_{\chi}$ with corresponding 3D boxes $P_{box} \in \mathbb{R}^{N \times C_{box}}$ from the instance adapter, the instance branch will spatially and temporally look back the image feature based on the referenced box coordinates and iteratively extract the abundant but more fine-grained representations to update pre-features. This branch can be roughly regarded as a multilayer Transformer-decoder-like \cite{transformer} module for 3D detection, which is briefly divided into two parts: box-level offset and embedding, as well as spatiotemporal sampling and fusion.

 \paragraph{Box-level Offset and Embedding} Different from the previous DETR-like 3D approaches such as DETR3D \cite{DETR3D21}, Polarformer \cite{PolarFormer22}, VEDet \cite{vedet} which iteratively refine only with 3D coordinate offset regression, the instance branch adopts the box-level offset regression based on $P_{box}$.  With this convenience, we can encode all the geometric-aware information of the entire box to substitute the transitional positional encoding, thereby expanding and enriching the space of feature expression rather than the superficial positional level. With it combined with the sparse instance features, there will be more geometric priors and implicit compensation in subsequent attention interactions. Precisely, we first categorize $P_{box}$ based on the element semantic of box dimension into four divisions, which are position $P_{pos} \in \mathbb{R}^{N \times 3}$ (\textit{i.e.}, $x, y, z$), scale $P_{sca} \in \mathbb{R}^{N \times 3}$ (\textit{i.e.}, $w, l, h$), velocity $P_{vel} \in \mathbb{R}^{N \times 2}$ (\textit{i.e.}, $v_x, v_y$), and orientation $P_{ori} \in \mathbb{R}^{N \times 2}$ (\textit{i.e.}, $sin(\theta_{yaw}), cos(\theta_{yaw})$) respectively. Then we introduce five separated linear projections $\{\textbf{E}_{l3}^j\}^{2}_{j=1} \in \mathbb{R}^{3 \times C}$, $\{\textbf{E}_{l2}^j\}^{2}_{j=1} \in \mathbb{R}^{2 \times C}$ and $\textbf{E}_g \in \mathbb{R}^{C \times C}$ for comprehensive encoding, of which the former four embed every category locally and the last one embeds them globally. The final box embedding $G_{\chi} \in \mathbb{R}^{N \times C}$ can be formulated by

 \begin{equation}
\label{box_embedding}
G_{\chi} = \textbf{E}_{g} \left[ \sum_{j = 1} ^2 \textbf{E}_{l3}^j \left(P_{d3}^j\right) + \sum_{j = 1} ^2 \textbf{E}_{l2}^{j} \left(P_{d2}^{j}\right) \right]
\end{equation}
where $P_{d3}^j, P_{d2}^j$ denote the three and two-dimensional categorized element respectively.

\paragraph{Spatiotemporal Sampling and Fusion} The sparse feature $F_{\chi}$ with the box embedding $G_{\chi}$ will be updated by the spatial and temporal sampling after being fed into the multi-headed self-attention block \cite{transformer}. Given the corresponding 3D coordinate $P_{\chi} \in \mathbb{R}^{N\times 3}$ from $P_{box}$, we first warp it from the 3D system to the 2D correspondence $p_{\chi} \in \mathbb{R}^{N\times 2}$ at the current time by the intrinsic and extrinsic matrix.

On the spatial hand, intending to access the target region, we sample the original feature to intermediately regress the existing offset from $p_{\chi}$ to the target. To expand the search breadth, we extend the sampling points analogous to Eq. \ref{bev_deformable} and enlarge the proportion of residual addition with a weight $\eta$. On the other temporal hand, with time going, there exists ego-car motion and object motion in the autonomous driving scenario, which requires compensation before sampling. In light of the short term in this sparse temporal stereo, \textit{i.e.}, $T_{\chi}$ is a small positive integer, we approximate the object motion as a uniform rectilinear motion. Thus, we first compensate $p_{\chi}$ with current 2D velocity $p_{vel} \in \mathbb{R}^{N\times 2}$, \textit{i.e.}, the 2D correspondence of $P_{vel}$ and then warp it into every coordinate system as $\{p_{\chi}^t\}_{t=1}^{T_{\chi}}$ of per historical time by the medium transition in the global world coordinate system. 
Then the per-frame sampled feature $F_{\delta t}, t \in \{0, 1, ..., T_{\chi}\}$ is formulated by
\vspace{-0.5em}
\begin{equation}
\begin{aligned}
\label{sampling}
    F_{\delta t} = \textbf{W}_{\chi} \sum_{k = 1} ^K A_{kt} \cdot &\textbf{W}^{'}_{\chi} \Phi [  F_{img}, \\
    & (\textbf{M}_{t} (p_{\chi} - \tau t\cdot p_{vel}) + \Delta p_{\chi k t})] 
\end{aligned}
\end{equation}
where $\tau$ is the time interval between every two adjacent frames, and $\textbf{M}_t$ is the ego-motion transform matrix from current time to previous $t$ time. Then the multi-frame features are fed into the sparse temporal encoder $f_{enc}$, a naive three-layer MLP, for temporal iterative fusion. Based on our approximation, the projection error will increase with $t$ larger. Thus we let $\lambda$ denote a constant in the range [0, 1], which is introduced for long-term suppression. With iterative fusion $F_{\delta (t-1)} \leftarrow f_{enc}(\{F_{\delta (t-1)}, \lambda F_{\delta t}\})$, we can get the final sparse sampled feature $F_{\delta}$ from $\{F_{\delta t}\}_{t=0}^{T_{\chi}}$. The whole box-level offset $\delta_{\chi}$ can be calculated as follows:
\begin{equation}
    \label{box_offset}
    F_{\chi} \leftarrow F_{\chi} + \eta F_{\delta}~~~~~~~\delta_{\chi} = f_{reg} \left\{F_{\chi} + G_{\chi} \right\}
\end{equation}
here $f_{reg}$ is the box offset regression function of every layer. Notably, we omit the calculation in the scale and view level of $F_{img}$ for simplicity.
\section{Experiments and Analysis}
\label{experiments}

\begin{table*}[t]
    \small
    \begin{center}

		\caption{Comparison results of LSS-based and two-stage detectors on 3D detection on the nuScenes \texttt{val} set. \dag~denotes the performance without future frames for a fair comparison.} \label{table:comparison_to_sota_detection_val_set}
        \vspace{-0.1in}
        \resizebox{.95\linewidth}{!}{
		\begin{tabular}{l|c|c|c|c|c@{\hspace{1.0\tabcolsep}}c@{\hspace{1.0\tabcolsep}}c@{\hspace{1.0\tabcolsep}}c@{\hspace{1.0\tabcolsep}}c}
        \toprule
            \textbf{Method} & \textbf{Backbone} & \textbf{Image Size}  & \textbf{mAP}$\uparrow$  &\textbf{NDS}$\uparrow$  & \textbf{mATE}$\downarrow$ & \textbf{mASE}$\downarrow$   &\textbf{mAOE}$\downarrow$   &\textbf{mAVE}$\downarrow$   &\textbf{mAAE}$\downarrow$  \\
            \midrule
            % ResNet50 256 x 704
            BEVDet~\cite{huang2021bevdet} & ResNet50 & 256 $\times$ 704 & 0.283 & 0.350 & 0.773 & 0.288 & 0.698 & 0.864 & 0.291 \\ % BEVDet Tab. 7
            BEVDet4D~\cite{huang2022bevdet4d} & ResNet50 & 256 $\times$ 704 & 0.322 & 0.457 & 0.703 & 0.278 & 0.495 & 0.354 & 0.206 \\ 
            BEVDepth~\cite{BEVDepth23} & ResNet50 & 256 $\times$ 704 & 0.330 & 0.436 & 0.702 & 0.280 & 0.535 & 0.553 & 0.227 \\
            SimMOD~\cite{SimMod23} & ResNet50 & 800 $\times$ 1333 & 0.331 & 0.427 & 0.721 & 0.267 & 0.401 & 0.810 & 0.184 \\
            BEVStereo~\cite{BEVStereo23} & ResNet50 & 256 $\times$ 704 & 0.346 & 0.452 & 0.659 & 0.277 & 0.550 & 0.498 & 0.228 \\ %  
            AeDet~\cite{aedet23} & ResNet50 & 256 $\times$ 704 & 0.358 & 0.473 & 0.655 & 0.273 & 0.493 & 0.427 & 0.216 \\
            SA-BEV~\cite{SA-BEV} & ResNet50 & 256 $\times$ 704 & 0.370 & 0.488 & 0.660 & 0.269 & 0.470 & 0.353 & 0.218 \\
            FB-BEV~\cite{FB-BEV} & ResNet50 & 256 $\times$ 704 & 0.378 & 0.498 & 0.620 & 0.273 & 0.444 & 0.374 & 0.200 \\
            BEVFormer v2~\cite{BEVFormerV2} \dag & ResNet50 & 640 $\times$ 1600 & 0.388 & 0.498 & 0.679 & 0.276 & 0.417 & 0.403 & 0.189 \\
            SOLOFusion~\cite{SOLOFusion23} & ResNet50 & 256 $\times$ 704 & 0.406           & 0.497          & 0.609        & 0.284        & 0.650        & 0.315         & 0.204               \\ 
            \rowcolor[gray]{.90}
            LSSInst & ResNet50 & 256 $\times$ 704 & \textbf{0.422} &\textbf{0.514} & 0.620 & 0.277 & 0.516 & 0.360 & 0.202 \\
            \bottomrule
        \end{tabular}
     }
    \end{center}
\end{table*}

\begin{table*}[t]
\small
    \begin{center}
	\caption{Comparison results of LSS-based detectors on 3D detection on the nuScenes \texttt{test} set. TTA denotes test time augmentation strategy. }
        \label{table:comparison_to_sota_detection_test_set}
        % \vspace{-0.1in}
        \resizebox{.95\linewidth}{!}{
        \begin{tabular}{l|c|c|c|c|c|c@{\hspace{1.0\tabcolsep}}c@{\hspace{1.0\tabcolsep}}c@{\hspace{1.0\tabcolsep}}c@{\hspace{1.0\tabcolsep}}c}
        \toprule
        \textbf{Method} & \textbf{Backbone} & \textbf{Image Size} & \textbf{TTA} & \textbf{mAP}$\uparrow$  &\textbf{NDS}$\uparrow$  & \textbf{mATE}$\downarrow$ & \textbf{mASE}$\downarrow$   &\textbf{mAOE}$\downarrow$  &\textbf{mAVE}$\downarrow$   &\textbf{mAAE}$\downarrow$  \\
        \midrule
        BEVDet~\cite{huang2021bevdet} & V2-99 & 900 $\times$ 1600 & \ding{52} & 0.424 & 0.488 & 0.524 & 0.242 & 0.373 & 0.950 & 0.148 \\
        BEVerse~\cite{beverse} & Swin-B & 900 $\times$ 1600 & \ding{52} & 0.393 & 0.531 & 0.541 & 0.247 & 0.394 & 0.345 & 0.129 \\
        BEVDet4D~\cite{huang2022bevdet4d} & Swin-B & 900 $\times$ 1600 & \ding{52} & 0.451 & 0.569 & 0.511 & 0.241 & 0.386 & 0.301 & 0.121 \\
        OA-BEV~\cite{OA-BEV} & V2-99 & 900 $\times$ 1600 & \ding{52} & 0.494 & 0.575 & 0.571 & 0.256 & 0.377 & 0.385 & 0.132 \\
        BEVDistill~\cite{bevdistill} & ConvNeXt-B & 640 $\times$ 1600 & \ding{56} & 0.496 & 0.594 & 0.475 & 0.249 & 0.378 & 0.313 & 0.125 \\
        BEVDepth~\cite{BEVDepth23} & ConvNeXt-B & 640 $\times$ 1600 & \ding{56} & 0.520 & 0.609 & 0.445 & 0.243 & 0.352 & 0.347 & 0.127 \\
        BEVStereo~\cite{BEVStereo23} & V2-99 & 640 $\times$ 1600 & \ding{56} & 0.525 & 0.610 & 0.431 & 0.246 & 0.358 & 0.357 & 0.138 \\
        AeDet \cite{aedet23} & ConvNeXt-B & 640 $\times$ 1600 & \ding{52} & 0.531 & 0.620 & 0.439 & 0.247 & 0.344 & 0.292 & 0.130 \\
        TiG-BEV \cite{TiG-BEV} & ConvNeXt-B & 640 $\times$ 1600 & \ding{52} & 0.532 & 0.619 & 0.450 & 0.244 & 0.343 & 0.306 & 0.132 \\
        SA-BEV~\cite{SA-BEV} & V2-99 & 640 $\times$ 1600 & \ding{56} & 0.533 & 0.624 & 0.430 & 0.241 & 0.338 & 0.282 & 0.139 \\
        FB-BEV~\cite{FB-BEV} & V2-99 & 640 $\times$ 1600 & \ding{56} & 0.537 & 0.624 & 0.439 & 0.250 & 0.358 & 0.270 & 0.128 \\
        SOLOFusion~\cite{SOLOFusion23} & ConvNeXt-B & 640 $\times$ 1600 & \ding{56} & 0.540 & 0.619 & 0.453 & 0.257 & 0.376 & 0.276 & 0.148 \\
        \rowcolor[gray]{.90}
        LSSInst & ConvNeXt-B & 640 $\times$ 1600 & \ding{56} & \textbf{0.546} & \textbf{0.629} & 0.464 & 0.251 & 0.341 & 0.265 & 0.120 \\
        \bottomrule
    \end{tabular}
    }
    \end{center}
    \vspace{-10pt}
\end{table*}

\begin{table*}[ht]
\small
    \begin{center}
    \caption{Generalization and Geometric-wise Results of LSSInst compared with LSS-type Baselines. ($\ddag$ please refer to Footnote \ref{fn2}).}
        \label{table:Generalization}
        \vspace{-0.1in}
        \resizebox{.90\linewidth}{!}{
        \begin{tabular}
        {@{\hspace{.5\tabcolsep}}l@{\hspace{.5\tabcolsep}}|@{\hspace{.5\tabcolsep}}c@{\hspace{.5\tabcolsep}}c|c@{\hspace{1.0\tabcolsep}}c@{\hspace{1.0\tabcolsep}}c@{\hspace{1.0\tabcolsep}}|c@{\hspace{1.0\tabcolsep}}cc}
        \toprule
        Method   & mAP$\uparrow$  &NDS$\uparrow$  & mATE$\downarrow$ & mASE$\downarrow$   &mAOE$\downarrow$ & Param (M) & \makecell[c]{Training Cost\\(min/epoch)} & \makecell[c]{Inference Cost\\(sec/frame)}  \\
        \midrule
\textbf{BEVDet} \cite{huang2021bevdet}   & 0.260    & 0.319      & 0.830    & 0.292   & 0.758      &  55.7   & 14              & 0.04\\%0.044               \\
\textbf{LSSInst} with BEVDet    & \textbf{0.310}      & \textbf{0.367}      & \textbf{0.771}     & \textbf{0.285}     & \textbf{0.658}    &  64.0  & 18    & 0.05\\%0.051    \\
\midrule
\textbf{BEVDepth4D} \cite{BEVDepth23}  & 0.343      & 0.458   & 0.691   & 0.281    & 0.610     &    83.5     & 11    & 0.10 \\ %0.097 \\
\textbf{LSSInst} with BEVDepth4D & \textbf{0.365}      & \textbf{0.477}      & \textbf{0.671}     & \textbf{0.275}      & \textbf{0.492}    &  91.8  & 13 & 0.11\\%0.109      \\
\midrule
\textbf{BEVStereo} \cite{BEVStereo23}    & 0.348  & 0.463   & 0.675   & 0.278  & 0.577   & 92.0 & 7  & 0.19\\%0.186  \\
\textbf{LSSInst} with BEVStereo & \textbf{0.372} & \textbf{0.481}  & \textbf{0.658}   & \textbf{0.275}     & \textbf{0.492}   & 102.3 & 10 & 0.21\\%0.208  \\
\midrule
\textbf{SOLOFusion} \cite{SOLOFusion23} & 0.406  & 0.497   & 0.609   & 0.284 & 0.650  & 64.4  &  23   & 0.07\\%0.065       \\
\textbf{LSSInst} with SOLOFusion  & \textbf{0.422} &\textbf{0.514} & ~0.620$^\ddag$ & \textbf{0.277} & \textbf{0.516}  & 72.8 & 26 & 0.08\\%0.078 \\
        
        \bottomrule
    \end{tabular}
    }
    
    \end{center}
\end{table*}

\subsection{Experimental Settings}
\paragraph{Dataset} We conducted extensive experiments on the nuScenes 3D detection benchmark \cite{Caesar2020nuScenesAM}, a large-scale dataset in the autonomous driving scene. This benchmark consists of 1,000 autonomous driving scenes, with each scene spanning approximately 20 seconds. The dataset is divided into 850 scenes for training (\texttt{train}) or validation (\texttt{val}) purposes and 150 scenes for testing (\texttt{test}).
Each frame in the dataset contains six cameras capturing surrounding views, along with a LiDAR-generated point cloud. The dataset provides annotations for up to 1.4 million 3D bounding boxes across 10 different classes.

\paragraph{Implementation Details}
We implemented our network framework utilizing the open-source MMDetection3D \cite{mmdet3d} in PyTorch.
The learning rate, optimizer, and data augmentation methods used were the same as those in BEVDepth. By default, We set the image size to 256 $\times$ 704 and utilized ResNet50 \cite{ResNet16}, pretrained on ImageNet \cite{ImageNet09}, as the image backbone. The size of the BEV feature in all our experiments was set to 128 $\times$ 128. Here we set $k=6$, $T_{\chi}=3$, $\eta=3$. The feature dimension $C$ is 256 and the box dimension $C_{box}$ is 10. The perception ranges for the $X$ and $Y$ axis was [-51.2m, 51.2m], and the resolution of each BEV grid was 0.8m. The time interval $\tau$ is 0.5s, and long-term suppression $\lambda$ is 0.6.

\subsection{Benchmark Results}
We compared our approach with LSS-based and two-stage state-of-the-art methods on the nuScenes \texttt{val} and \texttt{test} sets. The main results are presented in Tab. \ref{table:comparison_to_sota_detection_val_set} and Tab. \ref{table:comparison_to_sota_detection_test_set} respectively. On the \texttt{val} set, we evaluated the performance of LSSInst against other models with the same setting and without the CBGS strategy and future frame usage. The results clearly showcased the superiority of LSSInst, as it outperformed the current LSS-based SOTA, SOLOFusion by a margin of 1.6\% in mAP and 1.7\% in NDS, and the current two-stage SOTA, BEVFormer v2 by a margin of 3.4\% in mAP and 1.6\% in NDS. On the \texttt{test} set, our LSSInst achieves an mAP of 54.6\% and an NDS of 62.9\% without any additional augmentation, outperforming all LSS-based methods. Such improvements demonstrate the effectiveness of our LSSInst for improving LSS-based BEV perception with instance representations. 

\subsection{Generalization Ability and Geometric-Wise Boost}
To demonstrate the generalization ability of our LSSInst method, we selected prominent LSS-based methods as the BEV branch of LSSInst.
The results are presented in Tab. \ref{table:Generalization}. The table reveals that our LSSInst achieves notable improvements in mAP and NDS compared to standalone BEV detectors at a minor cost.
In spite of the impressive detection enhancement with 2-5\% mAP and NDS, the corresponding costs increase by an acceptable margin. 
In particular, among all the methods, there is a significant improvement in mATE \footnote{\label{fn2} Actually in the mATE column of Tab. \ref{table:Generalization}, 0.620 mATE in 0.422 mAP also beats 0.609 mATE in 0.402 mAP, please see the experimental verification in 
Supplementary Material.}, mASE, and mAOE, indicating that LSSInst can exploit fine-grained pixel-level features and better enhance perceptual capability with aspects of translation, scale, and orientation, which are all relevant in geometric-wise perception. 

\begin{table}[t]
\small
\centering
\setlength{\tabcolsep}{1.5pt} 
\caption{The noise resistance results for robustness.}
\label{table:noise}
\begin{tabular}{c|c|cc|cc}%cc}
\toprule
     \textbf{Method} &\textbf{Noise} & \textbf{mAP}\%$\uparrow$  & \textbf{Attenu.}\%$\downarrow$ &\textbf{NDS}\%$\uparrow$  & \textbf{Attenu.}\%$\downarrow$     \\ 
\midrule
Baseline & \multirow{2}*{0} & 35.74   & -  &  46.84 &  - \\ 
LSSInst  & ~                & 38.28   & -  &  49.43 &  - \\ 
\midrule
Baseline & \multirow{2}*{0.5\%} & 35.38   & 1.01  &  46.44 &  0.85 \\ 
LSSInst  & ~                    & \textbf{38.01}   & 0.71  &  \textbf{49.19} &  0.49 \\ 
\midrule
Baseline & \multirow{2}*{1\%} & 34.13   & 4.5  &  45.31 &  3.3 \\ 
LSSInst  & ~                    & \textbf{36.58}   & 4.4  &  \textbf{47.85} &  3.2  \\ 
\midrule
Baseline & \multirow{2}*{2\%} & 30.32   & 15.2  &  41.77 &  10.8 \\ 
LSSInst  & ~                    & \textbf{32.23}   & 15.8  &  \textbf{44.28} &  10.4 \\ 
\midrule
Baseline & \multirow{2}*{3\%} & 26.22   & 26.7  &  37.84 &  19.2 \\ 
LSSInst  & ~                    & \textbf{27.92}   & 27.1  &  \textbf{40.47} &  18.1 \\ 
\bottomrule
    \end{tabular}
\end{table}

\subsection{Noise Resistance for Practical Robustness}
Although we have verified the high performance of LSSInst on nuScenes \cite{Caesar2020nuScenesAM}, even the large-scale autonomous driving dataset inevitably contain disturbances in the extrinsics obtained when sensors collect data in huge quantities. In actual autonomous driving scenarios, the detector is required to be resistant to the disturbance noise caused by small measurement errors. Therefore, here we add a set of random rotation noise with increasing proportions to the extrinsics, exploring the robustness of LSSInst under inaccurate extrinsics. Here, the baseline is BEVDepth4D \cite{BEVDepth23} with 4 frames. As shown in Tab. \ref{table:noise}, we demonstrate that LSSInst maintains good robustness, exhibiting higher performance and smaller overall attenuation.

\begin{table}[t]
\small
\centering
\caption{\fontsize{9bp}{10bp} Query Composition}
\setlength{\tabcolsep}{10pt} 
\renewcommand\arraystretch{1.1}
\begin{tabular}{@{}c|cc@{}}
\toprule
Composition of Queries & mAP$\uparrow$ & NDS$\uparrow$ \\
\midrule
450~$Q_\gamma$ & 0.157 & 0.226 \\
900~$Q_\gamma$ & 0.263 & 0.297 \\
450~$Q_\beta$ & 0.331 & 0.447 \\
900~$Q_\beta$ & 0.330 & 0.446 \\
450~$Q_\beta$ + 450~$Q_\gamma$ & \textbf{0.362} & \textbf{0.474} \\
\bottomrule
\end{tabular}
\label{table:abl-Q}
\end{table}

\begin{table}[ht]
\small
\centering
\caption{\fontsize{9bp}{10bp} Segmentation mIoU}
\setlength{\tabcolsep}{12pt} 
\begin{tabular}{@{}l|cc@{}}
\toprule
Methods & with GT & with baseline \\ \midrule
Baseline & 44.56 & - \\ \midrule
LSSInst & \textbf{46.63} & \textbf{66.21} ($>$50) \\
\bottomrule
\end{tabular}
\label{table:abl-mIoU}
\end{table}

\subsection{Multiplicate Queries Ablations}
\label{abalation}

To further investigate the impact of the multiplicate queries, as shown in Tab. \ref{table:abl-Q}, we explored two scenarios: using only proposal queries (referred to as $Q_\beta$) or learnable potential queries (referred to as $Q_\gamma$), and incorporating both queries. Here we follow the classical setting of query maximum number by default 900 in \cite{DETR3D21}.
We can observe that on the one hand, relying solely on the potential queries cannot play a major role, and even utilizing all 900 queries yielded mediocre performance, 
which shows the slow convergence because of the initialization semantic dispersity without the scene-level information basis from BEV as aforementioned.
On the other hand, though the proposal queries from BEV alone can achieve overall good results, adding more queries does not achieve a better improvement, 
which proves its overfitting characteristics for the scene and the fact of the neglectful detection of missing objects in the scene.
However, when incorporating two kinds of queries, the performance is further improved and reaches a new level. 
It can be concluded that these two types of queries play their unique roles, and their inseparable and complementary synergy enables the model to have a comprehensive understanding from the global scene level to the local instance level.

\subsection{BEV-to-Instance Semantic Coherence} To confirm the BEV-to-instance semantic coherence, we conduct the relevant experiments in two aspects. Assuming there is only one foreground class, we calculate the mIoU metric of semantic segmentation compared with the ground truth and baseline as shown in Tab. \ref{table:abl-mIoU}. According to the results with the ground truth, LSSInst is observed as having better semantic maintenance than the LSS baseline, which shows the improvement of perceptual capability to the extra BEV-insensitive objects in the scene. When it comes to the mIoU with the LSS baseline, the value 66.21\% is over 50\% which also indicates the promising BEV-to-instance semantic coherence. Notably, the relevant qualitative results can be referred to in Supplementary Material.

\section{Conclusion}
Existing LSS-based methods make efforts to build up a desirable BEV representation, but they ignore its inherent shortage of geometric loss in the formulation, suppressing its generality in 3D perception. In this paper, we propose LSSInst, a two-stage detector that improves the geometric modeling of the BEV perception with instance representation. To address the challenge of the gap between two representation spaces, we propose the instance adaptor to keep the BEV-to-instance semantic coherence. Then a newly-designed instance branch is introduced to look back for fine-grained geometric matching and feature aggregation. Extensive experimental results demonstrated that our framework is of great generalization ability in modern LSS-based BEV perceptions and excellent performance, surpassing the current state-of-the-art works. We hope that our work will inspire further exploration of generalized 3D perception in more complex and fine-grained outdoor-scene tasks.

\section*{Acknowledgement}
This work is partially supported by the National Key R\&D Program of China (No. 2022ZD0160101) and the National Natural Science Foundation of China (No. 62206244).

{
    \small
    \bibliographystyle{ieeenat_fullname}
    \bibliography{main}
}
\clearpage
\setcounter{page}{1}
\setcounter{section}{0}
\renewcommand\thesection{\Alph{section}}
\maketitlesupplementary

\noindent The supplementary document is organized as follows:

\begin{itemize}
\item Sec. \ref{GeoPorj} depicts the 3D geometric projection details of the instance branch.
\item Sec. \ref{NetArch} provides the detailed module network architectures and design rationality.
\item Sec. \ref{PropGen} describes the generation details of BEV proposals.
\item Sec. \ref{ExpSetsExts} provides extensive experimental details and additions of LSSInst.
\item Sec. \ref{QualiRes} provide the qualitative results and visualization analysis.

\end{itemize}

\section{Instance-level 3D Geometric Projection}
\label{GeoPorj}

For the 3D position ego coordinates $P_{pos} \in \mathbb{R}^{N \times 3}$ at the current time, below are the detailed multi-view geometric projection for instance-level representations. Firstly, on the \textbf{spatial} hand, $P_{pos}$, \textit{i.e.}, $(x, y, z)$ is warped into the camera coordinate system by using the per-view extrinsics $\textbf{M}_{cam} = [ \mathbf{R} | \mathbf{t} ] \in $ SE3 and intrinsics as 2D points $p_{\chi}$, \textit{i.e.}, $(u,v)$ as follows: 
\begin{equation}
    \label{projection}
    \left[\begin{array}{c}
    x_c\\
    y_c\\
    z_c\\
    1
    \end{array}\right]=
    \left[\begin{array}{cc}
    \mathbf{R} & \mathbf{t} \\
    0^3 & 1
    \end{array}\right]
    \left[\begin{array}{c}
    x\\
    y\\
    z\\
    1
    \end{array}\right]
\end{equation}
\begin{equation}
    \quad
    z_c\left[\begin{array}{c}
    u\\
    v\\
    1
    \end{array}\right]=
    \left[\begin{array}{ccc}
    f_{x} & 0 & c_{x} \\
    0 & f_{y} & c_{y} \\
    0 & 0 & 1
    \end{array}\right]
    \left[\begin{array}{c}
    x_c\\
    y_c\\
    z_c
    \end{array}\right]
\end{equation}
where $x_c, y_c, z_c$ are the coordinates in camera system, $f_{x},f_{y},c_{x},c_{y}$ are camera intrinsic parameters, and $\mathbf{R} \in \mathbb{R}^{3\times3}, \mathbf{t} \in \mathbb{R}^{3\times1}$ denote the spatial rotation and translation matrices. 

Secondly, on the \textbf{temporal} hand, we warp $p_{\chi}$ to the target coordinate system at time $t$, and for the unified expression, the set of target systems includes the current system, \textit{i.e.}, $0 \in \{t\}$. On the basis of Sec. 
% \ref{SparseBranch} 
3.3 and Eqn. 
% \ref{sampling}
5, below is the detailed formulation of $\textbf{M}_{t}$. Given both extrinsic calibration matrices to the world coordinate system $\textbf{M}_{cur2w}, \textbf{M}_{tgt2w} \in $ SE3, we can construct the transformation matrix $\textbf{M}_{t}$ from the current system to the target one by
\begin{equation}
    \label{temproal_transformation}
    \textbf{M}_{t} = \textbf{M}_{tgt2w}^{-1} \times \textbf{M}_{cur2w} =
    \left[\begin{array}{cc}
    \mathbf{R}_{tgt} & \mathbf{t}_{tgt} \\
    0^3 & 1
    \end{array}\right]
\end{equation}
where $\mathbf{R}_{tgt} \in \mathbb{R}^{3\times3}, \mathbf{t}_{tgt} \in \mathbb{R}^{3\times1}$ denote the overall temporal rotation and translation matrices. 

\section{Network Architectures}
\label{NetArch}
\paragraph{Feature Converter}
The detailed architecture of the feature converter module is the combination of a $3\times3$ kernel-size convolution layer with $1$ padding and batch normalization, aiming to learn an inter-space adaptation from resampled BEV feature to sparse instance features. Here we convert the whole BEV feature into the adaptive space at first in practice for implementation convenience, and we give a short proof to show the equivalence as follows:

\begin{equation*}
\begin{aligned}
     \widehat{F}_{\alpha} &= f_{cnv} \left (\{F_{\alpha}^i\}^{N_{\beta}}_{i=1} \right ) && \qquad \text{(Eqn. 
     % \ref{query_composition}
     3)}\\
     &=  f_{cnv} \left ( \left \{ \Phi \left[F_{\beta}, P_{\beta}^i \right] \right\}_{i=1}^{N_{\beta}} \right ) && \qquad \text{(Eqn. 
     % \ref{bev_deformable}
     2)}\\
     &=  f_{cnv} \left ( \Phi \left [F_{\beta}, \left \{P_{\beta}^i\right\}_{i=1}^{N_{\beta}} \right ] \right )  \\
    &=  \Phi \left[f_{cnv} \left (F_{\beta}\right), P_{\beta}  \right ]
\end{aligned}
\end{equation*}
where $\widehat{F}_{\alpha}$ denotes the converted feature from the set of $F_{\alpha}^i$, and we omit the specific resampling multipliers in Eqn. 
% \ref{bev_deformable} 
2 for simplicity.

\paragraph{Sparse Temporal Encoder} The specific architecture of the sparse temporal encoder is a naive three-layer MLP with GeLU \citep{GeLU} for sparse temporal fusion from $2C$ to $C$. Below are the detailed procedures in Alg. \ref{alg}. As shown in the algorithm, when the iterative fusion ends, the accumulated highest order of $\lambda$ will come to $(t-1)$, \textit{i.e.}, the impact equals to a $\lambda^{t-1}$ multiplier for every $F_{\delta t}$, which indeed acts as the desirable long-term suppression.

\begin{algorithm}
\caption{The pseudo-code of sparse temporal fusion}\label{alg}
\begin{algorithmic}[1]
\Require $T_{\chi} \in \mathbb{N}^{+},~ 0 < T_{\chi} \leq T,~ 0 < \lambda < 1$
% \Ensure $y = x^n$
\State $t \gets T_{\chi}$
\While{$t \neq 0$}
    \State $F_{\delta t} \gets \lambda F_{\delta t}$ \Comment{$F_{\delta t}$ is formulated by Eqn. %\ref{sampling}
    5}
    \State $F_{\delta (t-1)} \gets concat[F_{\delta (t-1)}, F_{\delta t}]$
    \State $F_{\delta (t-1)} \gets f_{enc}(F_{\delta (t-1)})$
    \State $t \gets t - 1$
\EndWhile
\end{algorithmic}
\end{algorithm}

\begin{table*}[t]
    \small
    \begin{center}
		\caption{Comparison results of LSS-based detectors for 3D detection on the nuScenes \texttt{val} set.  All methods in the table are trained with CBGS.} \label{table:cbgs_comparison}
        \resizebox{1.0\linewidth}{!}{
		\begin{tabular}{l|c|c|c|c|c@{\hspace{1.0\tabcolsep}}c@{\hspace{1.0\tabcolsep}}c@{\hspace{1.0\tabcolsep}}c@{\hspace{1.0\tabcolsep}}c}
        \toprule
            \textbf{Method} & \textbf{Backbone} & \textbf{Image Size}  & \textbf{mAP}$\uparrow$  &\textbf{NDS}$\uparrow$  & \textbf{mATE}$\downarrow$ & \textbf{mASE}$\downarrow$   &\textbf{mAOE}$\downarrow$   &\textbf{mAVE}$\downarrow$   &\textbf{mAAE}$\downarrow$  \\
            \midrule
            % ResNet50 256 x 704
            BEVDet~\citep{huang2021bevdet} & ResNet50 & 256 $\times$ 704 & 0.298 & 0.379 & 0.725 & 0.279 & 0.589 & 0.860 & 0.245 \\ % BEVDet Tab. 7
            BEVDet4D~\citep{huang2022bevdet4d} & ResNet50 & 256 $\times$ 704 & 0.322 & 0.457 & 0.703 & 0.278 & 0.495 & 0.354 & 0.206 \\ 
            BEVDepth~\citep{BEVDepth23} & ResNet50 & 256 $\times$ 704 & 0.351 & 0.475 & 0.639 & 0.267 & 0.479 & 0.428 & 0.198 \\
            STS~\citep{STS22} & ResNet50 & 256 $\times$ 704 & 0.377 & 0.489 & 0.601 & 0.275 & 0.450 & 0.446 & 0.212 \\
            BEVStereo~\citep{BEVStereo23} & ResNet50 & 256 $\times$ 704 & 0.372 & 0.500 & 0.598 & 0.270 & 0.438 & 0.367 & 0.190 \\ %  
            AeDet~\citep{aedet23} & ResNet50 & 256 $\times$ 704 & 0.387 & 0.501 & 0.598 & 0.276 & 0.461 & 0.392 & 0.196 \\
            SA-BEV~\citep{SA-BEV} & ResNet50 & 256 $\times$ 704 & 0.386 & 0.512 & 0.612 & 0.266 & 0.351 & 0.382 & 0.200 \\
            SOLOFusion~\citep{SOLOFusion23} & ResNet50 & 256 $\times$ 704 & 0.427 & 0.534 & 0.567 & 0.274 & 0.511 & 0.252 & 0.188 \\
            \rowcolor[gray]{.90}
            LSSInst & ResNet50 & 256 $\times$ 704 & \textbf{0.429} & \textbf{0.537} & 0.595 & 0.281 & 0.423 & 0.273 & 0.202 \\
            \bottomrule
        \end{tabular}
     }
    \end{center}
\end{table*}

\section{Proposal Generation}
\label{PropGen}
We describe the generation pipeline for BEV proposals from the proposal head in this section. The proposal head can be a very lightweight BEV detection head, like CenterHead \citep{centerpoint}, only for generating the raw BEV proposals with their scores $\{\rho^i, s^i\}_i$. With the non-maximum suppression (NMS) operation with a score threshold, we can obtain the 3D bounding box candidates $C_o$. Here the threshold is set as 0.1. Considering the variable amount of candidates, we re-filter them by top-k as follows, and here k is set as 450, half of the classical total number of 3D queries.
\begin{equation}
\label{NMS}
C_o := \text{top-}k \left (\text{NMS}\left [\{\rho^i, s^i\}_i\right] \right )
\end{equation}

Notably, there also exists the possibility that the amount is smaller than $k$. We add the blank padding for the rest, where the position is random with a $\pi/2$ yaw, and both scale and velocity are zero.

\section{Experimental Settings and Extensions}
\label{ExpSetsExts}

\paragraph{Evaluation Metrics}
For 3D object detection in the nuSense benchmark, our study utilizes a set of official predefined metrics to evaluate the performance of our approach. These metrics include mean Average Precision (mAP), Average Translation Error (ATE), Average Scale Error (ASE), Average Orientation Error (AOE), Average Velocity Error (AVE), Average Attribute Error (AAE), and nuScenes Detection Score (NDS). Different from direct 3D IoU usage, here mAP is based on the BEV center distance and is calculated by averaging over distance thresholds of 0.5m, 1m, 2m, and 4m for ten different classes of objects, including car, truck, bus, trailer, construction vehicle, pedestrian, motorcycle, bicycle, barrier, and traffic cone. In addition to mAP, NDS is a comprehensive metric that takes into account other indicators to assess the overall detection performance. The remaining metrics are designed to measure the precision of positive results in concerned aspects, such as translation, scale, orientation, velocity, and attribute.

\subsection{Experimental Settings}

Our implementation is conducted in MMDetection3D \citep{mmdet3d} with one NVIDIA A100 40G GPU node. The adoption of data augmentation strategies follows the setting of the BEV branch. Specifically, the augmentation strategies can be random flips along the X and Y axes, random scaling and rotation in a limited range in the image or BEV level. As for the FPN \citep{FPN} before each branch, we follow the settings of BEVDepth \citep{BEVDepth23} and DETR3D \citep{DETR3D21}, respectively and choose SECONDFPN \citep{SECOND18} with 128-dimensional output and standard FPN \citep{FPN} with 256-dimensional output. We select AdamW \citep{adamw} as the optimizer and set the learning rate as 2e-4. Notably, in the ablation study, we selected BEVDepth \citep{BEVDepth23} as the BEV branch in the ablation baseline for convenient experimental conduction. Here the BEV branch used 1+2 frames and the sparse branch of the ablation baseline didn't use temporal information except for the frame ablation.

\subsection{Experimental Extensions}
In this section, we conducted the experimental extensions to show more persuasive performance results and ablation. 

\subsubsection{Performance Extension}
CBGS strategy as an incremental trick is popular in several works to further increase model performance. In order to further compare with the LSS-based state-of-the-art methods trained with CBGS, we conducted a performance evaluation in the nuScenes val set. As shown in Tab. \ref{table:cbgs_comparison}, our LSSInst achieves an mAP of 42.9\% and an NDS score of 53.8\%, outperforming all existing methods. These results further demonstrate the missing details improvement and inherent effectiveness of our method despite the class imbalance compensation using CBGS.

\subsubsection{Verification for Translation Improvement}
The mA*E is designed to measure a property (here we use * to denote this) by the mean statistical error. It should be pointed that it's actually based on the predicted instances, which does not consider the confidence threshold. In other words, the comparison of mean average errors mA*E is more persuasive and meaningful based on the same AP which indicates the same level of detecting boxes according to the calculation formulas \cite{Caesar2020nuScenesAM}. Considering that we introduce more queries to capture the missing objects, it also means we are more likely to yield lower mATE for those low-score predictions. In practice, we enhance the confidence level and decrease the output box number to show the mATE variation as shown in the table below. When we change to the 100 output number, we can easily observe the better mATE as well as higher mAP than the baseline \cite{SOLOFusion23}.

\begin{table}[t]
\label{mATE}
\centering
\caption{The experimental verification of mATE improvement.}
\vspace{0.1in}
\begin{tabular}{l|c|c|c}
\toprule

\multicolumn{2}{c|}{\textbf{Method}}  & \multirow{2}{*}{\textbf{mAP}$\uparrow$} & \multirow{2}{*}{\textbf{mATE}$\downarrow$} \\ \cline{1-2}
\multirow{6}{*}{LSSInst} & Out\_Box\_Num &         \\ \cline{2-4}
                         & 300 & 42.2    & 0.620   \\ 
                         & 250 & 42.2    & 0.619   \\
                         & 200 & 42.1    & 0.617   \\
                         & 150 & 42.0    & 0.614   \\
                         & 100 & \textbf{41.4}    & \textbf{0.608}   \\
\bottomrule
\multicolumn{2}{l|}{SOLOFusion \citep{SOLOFusion23}} & 40.6    & 0.609   \\
\bottomrule
\end{tabular}
\end{table}

\begin{figure}[t]
  \centering
  \includegraphics[ width=\linewidth]{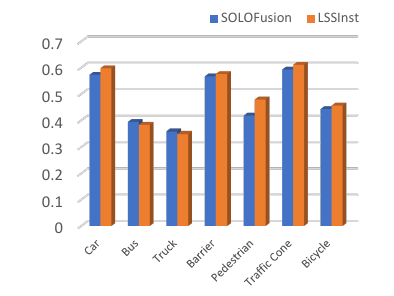}
  \vspace{-0.1in}
  \caption{Comparison results of per-classes mAP on nuScenes $\mathtt{val}$ set.}
  \label{per-class}
\end{figure}

\subsubsection{Results of Category-level Improvement}
This section shows the per-class comparison results between SOLOFusion and LSSInst on the nuScene $\mathtt{val}$ and $\mathtt{test}$ set. As illustrated in Fig. \ref{per-class}, we can observe the BEV-insensitive categories such as the traffic cone and bicycle, especially pedestrian have been detected with a remarkable margin. It's favorable for the improvement of the classes with variable movements or dispersed locations since there is a large proportion of human beings (pedestrians) in the auto-driving scenario.

\begin{table}[t]
\small
\centering
\caption{\fontsize{9bp}{10bp} Box-level Embedding}
\setlength{\tabcolsep}{10pt} 
\renewcommand\arraystretch{1.1}
\begin{tabular}{@{}ccc|cc@{}}
\toprule
Center & Box & Box w/ BE & mAP$\uparrow$ & NDS$\uparrow$ \\
\midrule
& & & 0.343 & 0.458 \\
\checkmark & & & 0.354 & 0.467 \\
& \checkmark & & 0.354 & 0.466 \\
& & \checkmark & \textbf{0.362} & \textbf{0.474} \\
\bottomrule
\end{tabular}
\label{table:abl-be}
\end{table}

\subsubsection{Ablation Extension of Whistles and Bells}

The ablation study below reveals the role and function of each component in our framework. Notably, we select BEVDepth as the BEV branch in the ablation baseline for convenient experimental conduction. The sparse branch of the ablation baseline does not use temporal information except for the frame ablation.

\paragraph{Box-level embedding}
In order to showcase the impact of box-level embedding, we conducted the ablation experiment, and the results are presented in Tab. \ref{table:abl-be}. In this experiment, we compared different approaches: utilizing only the center points (referred to as \textbf{Center}) or the bounding boxes (\textbf{Box}) predicted in the BEV detection and incorporating the bounding boxes along with their corresponding box embedding (\textbf{BE}). We can both find the same increase by a margin, which indicates that there is no difference between the two types of offset regression, excluding the possibility of using Box to bring additional information compared with Center. However, by incorporating the box-level embedding, we observed a further remarkable improvement over center point inheritance alone. This significant improvement clearly demonstrates the encoding of candidate boxes helps enhance the geometric priors of the queries, thereby improving the extraction of detailed object features from the image. This compensates for the limitations of the BEV representation and enables a more comprehensive understanding of instances.

\paragraph{Offline Temporal Sampling} Here we particularly change the frame from 3 to 4 for a more comprehensive observing range. As shown in Tab. \ref{table:4f}, the results reveal a fluctuating trend. The performance improves gradually as the number of frames increases up to 3, but when the number of frames reaches 4, the performance starts to decline, reflecting a bottleneck. 
This observation not only indicates that our geometric-guided temporal fusion is helpful for short-term matching and alignment but also shows the theoretical long-term error and verifies the limited approximation mentioned in Sec. 
% \ref{SparseBranch}
3.3 even though adding the suppression. It can be inferred that as the look-back window increases longer, the objects have moved a larger distance within the interval of much more than 3$\sim$4$\tau$ = 1.5$\sim$2 seconds, and the variable movement makes it challenging to align the features under short-term geometric constraints, leading to a continuous decrease in performance. In the future, we will adopt online temporal sampling to acquire a wider temporal range to improve the problem.

\begin{table}[t]
\small
\centering
\setlength{\tabcolsep}{5pt}
\caption{The ablation results of frame-level extensions}
        \label{table:4f}
        
        \begin{tabular}{c|cc|ccc}%cc}
\toprule
     \textbf{Frame} & \textbf{mAP}$\uparrow$  &\textbf{NDS}$\uparrow$  & \textbf{mATE}$\downarrow$ & \textbf{mASE}$\downarrow$   &\textbf{mAOE}$\downarrow$ \\ % &\textbf{mAVE}$\downarrow$   &\textbf{mAAE}$\downarrow$  \\
    \midrule
   BEV Only & 0.366         & 0.477   & 0.661  &  0.278 &  0.625 \\ % &  0.327  &  0.205         \\
   % \midrule
1&0.381          & 0.494   & 0.662  &  0.271 &  0.473 \\ % &  0.358  &  0.207       \\
2&0.382          & 0.495  & 0.654  &  0.271 &  0.470 \\ % &  0.360  &  0.207        \\
3&\textbf{0.389} & \textbf{0.497} & \textbf{0.652}  &  \textbf{0.270} &  \textbf{0.454} \\ % &  0.378  &  0.223\\
4&0.383          & 0.496  & 0.659  &  0.273 &  0.462 \\ % &  0.366  &  0.198\\
        \bottomrule
    \end{tabular}
\end{table}

\begin{table}[t]
\small
\centering
\setlength{\tabcolsep}{10pt}
\caption{Adaptor Ablation}
    \begin{tabular}[h]{cc|cc}
\toprule
     BDR       &FC          & mAP$\uparrow$ & NDS$\uparrow$   \\
    \midrule
   	&	&		0.3623 	&	0.4741 \\
\checkmark	&	&	0.3647 	&	0.4769 \\

& \checkmark	&	0.3651 	&	0.4753 \\
 % \rowcolor[gray]{.90}
\checkmark	& \checkmark	& \textbf{0.3661} & \textbf{0.4779} \\

\bottomrule
    \end{tabular}
    % }
  \label{table:abl-da}
\end{table}

\begin{table}[t]
\small
    \parbox{.48\linewidth}{
\centering
    \caption{\fontsize{9bp}{10bp}Point Ablation}
    \label{table:abl-p}
    \begin{tabular}[h]{c|cc}
\toprule
    Points       & mAP$\uparrow$ & NDS$\uparrow$   \\
    \midrule
   	1   &	0.365	&	0.477  \\
        2	&	\textbf{0.369}  	&	0.478  \\
	4	&	0.361 	&	0.472 \\
        6	&	0.364 	&	\textbf{0.479}\\
\bottomrule
    \end{tabular}
    % }
  
    }
    \hfill
    \parbox{.48\linewidth}{
\centering
    \caption{\fontsize{9bp}{10bp}Weight Ablation}
    
    \begin{tabular}[h]{c|cc}
\toprule
    Weight       & mAP$\uparrow$ & NDS$\uparrow$   \\
    \midrule
   	1   &	0.365	&	0.477  \\
        2	&	\textbf{0.370}  	&	0.478  \\
	3	&	0.366 	&	\textbf{0.480} \\
        4	&	0.362 	&	0.474\\
\bottomrule
    \end{tabular}
  \label{table:abl-w}
    }
\end{table}

\paragraph{Instance Adaptor} 
To showcase the effectiveness of the instance adaptor module in LSSInst, we conducted a series of ablation experiments, as depicted in Tab. \ref{table:abl-da}. In this table, \textbf{BDR} denotes the BEV feature deformable resampling, and \textbf{FC} represents the feature converter. The results indicate that both sub-modules achieved a 0.2\% improvement in both mAP and NDS compared to the baseline. When combined, they contributed to a total improvement of 0.4\%. This indicates that our instance adaptor module effectively preserves the semantic coherence between BEV and instance representations, enabling effective improvement of BEV features using instance-level information.

\paragraph{Spatial Sampling and Fusion} As for spatial sampling, we utilize deformable attention to aggregate features from multiple sampling points. As shown in Tab. \ref{table:abl-p}, when we increase the number of sampling points to 2, there is a 0.4\% improvement in mAP, indicating that richer spatial sampling helps enrich features and optimize intermediate representations. However, further increasing the number of sampling points results in a performance decline, which may be owing to the smaller resolution of the feature map.
As shown in Tab. \ref{table:abl-w}, We explore the performance of different weights assigned to image features. The results reveal that increasing the weight of image features to 2 leads to a 0.5\% improvement in mAP. This indicates that increasing the weight of image features during spatiotemporal sampling helps enhance the representation ability of queries. The network tends to utilize a larger weight of image features, which further verifies the effectiveness of our designed instance branch for improving intermediate representations.

\begin{figure}[t]
  \centering
  \includegraphics[width=0.9\linewidth]{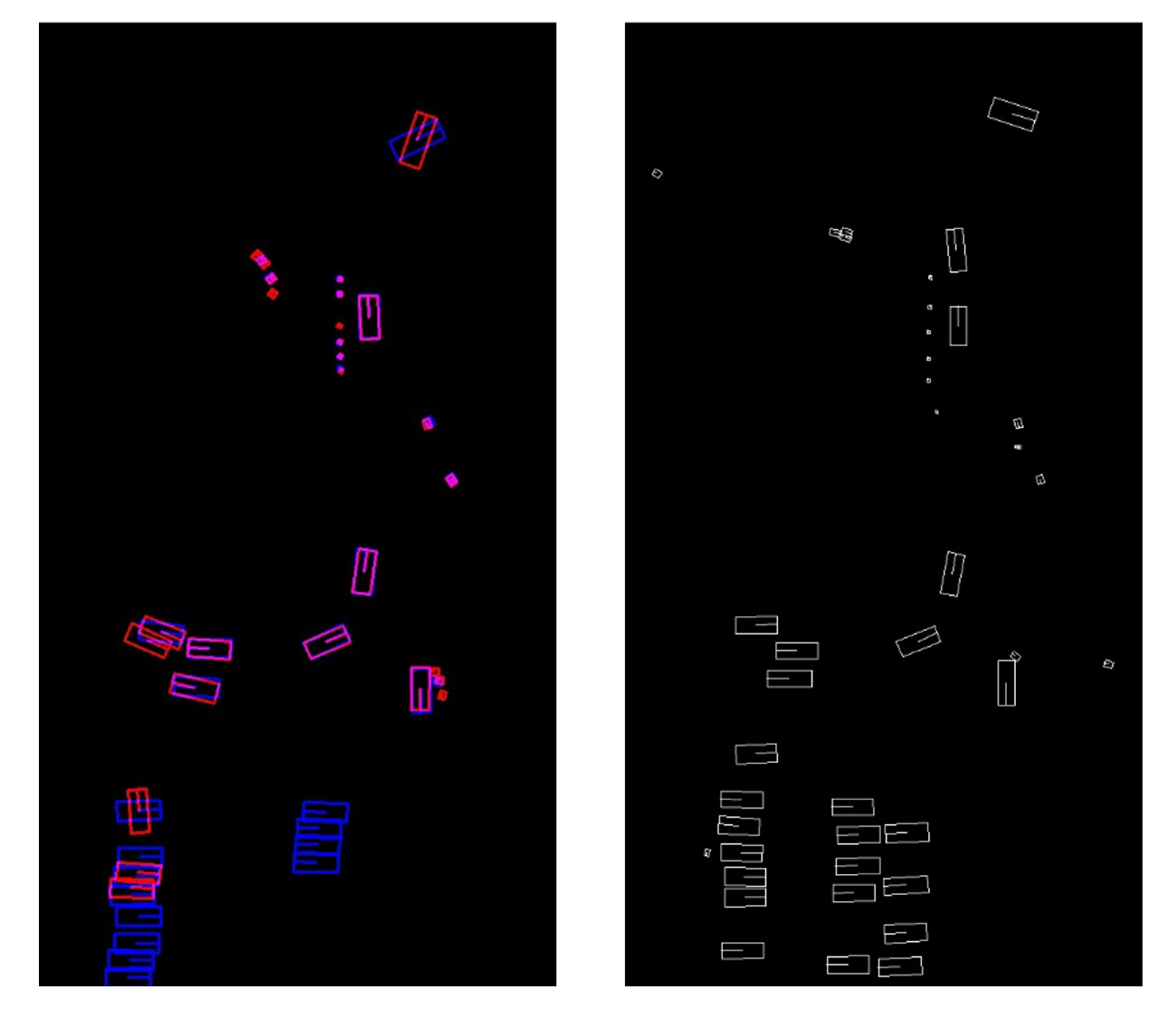}
  % \vspace{-0.1in}
  \caption{Qualitative comparison between baseline proposals (red), predictions (blue), their superposition (purple), and GT (white).}
  \label{Coherence}
\end{figure}

\begin{figure*}[t]
  \centering
  \includegraphics[width=0.9\linewidth]{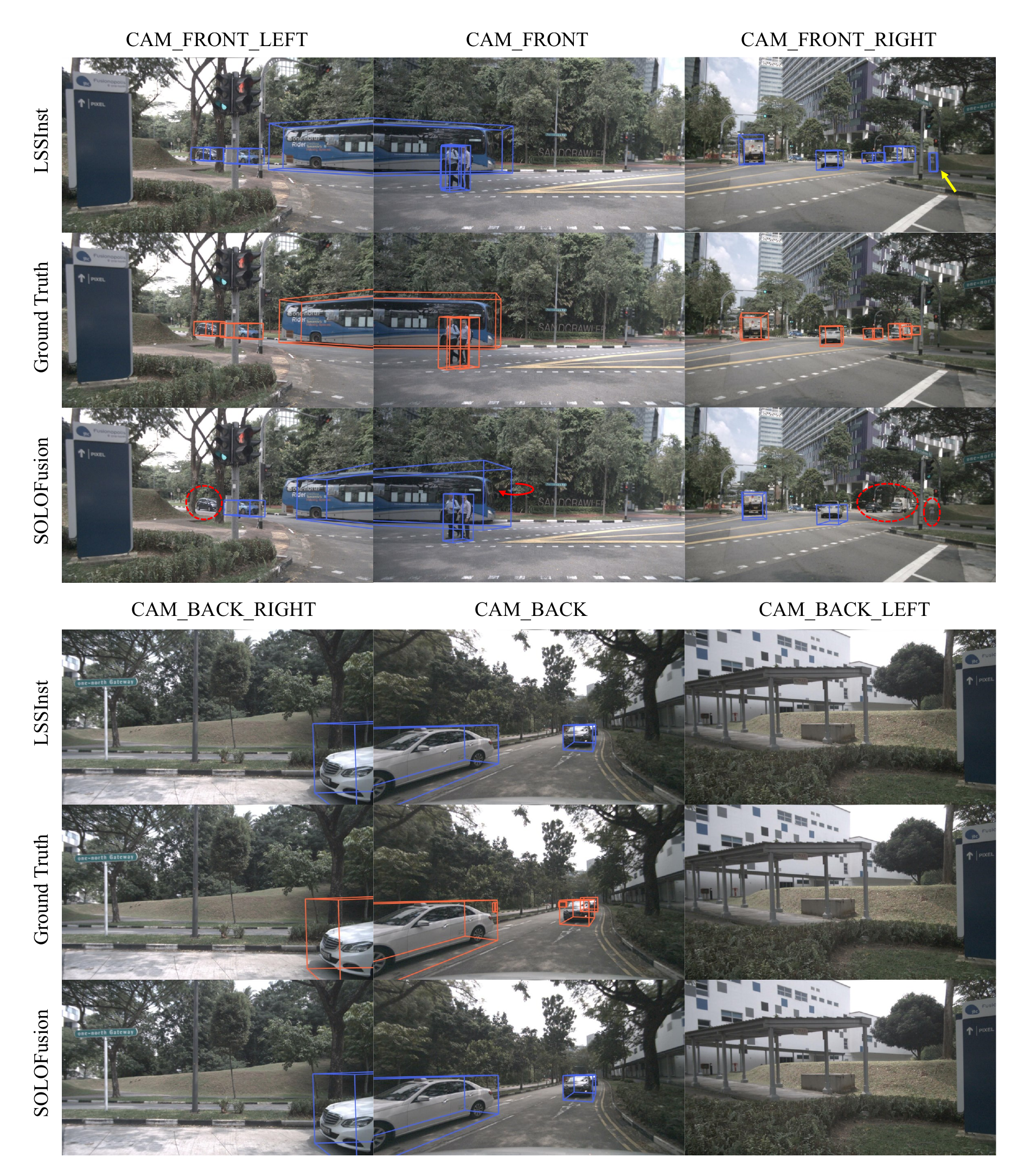}
  % \vspace{-0.1in}
  \caption{Comparison of LSSInst, the ground truth, and SOLOFusion on nuScenes $\mathtt{val}$ set.}
  \label{visual1}
\end{figure*}

\section{Qualitative Results}
\label{QualiRes}

\subsection{Qualitative comparison of BEV-to-Instance Coherence}
Despite the semantic segmentation mIoU result between LSSInst and the baseline is 66.21\% which indicates that our method possesses a desirable semantic scene-layout basis and keeps better semantic coherence, to illustrate this point apparently, here visualize the comparison results between proposals and predictions. It can be more clearly observed not only the coherence but also extra improvement on the basis. As shown in Fig. \ref{Coherence}, where blue is yielded by BEVInst, red denotes the proposals, purple means their superposition, and white means GT. We can first conclude that purple boxes occupy the majority. Then there are many red boxes for false or missed detection and some blue boxes for orientation correction or additional detection which match the white boxes much more, which directly proves the improvement. 

\subsection{Visualization}
In this section, we show the visualization comparison results for 3D object detection among LSSInst, ground truth, and current SOTA method SOLOFusion. As shown in Fig. \ref{visual1}, LSSInst has a higher recall and detects more inapparent and occluded objects. For example, our model successfully detects distant cars and trucks in the $\mathtt{CAM\_FRONT\_LEFT}$ and $\mathtt{CAM\_FRONT\_RIGHT}$ views, especially the vehicle occluded by trees and the inapparent car with dark color which is highly similar with the background. Significantly, as the yellow arrow shown in the $\mathtt{CAM\_FRONT\_RIGHT}$ view, we surprisingly find the pedestrian, who is so tiny and indistinct that he/she is even ignored by the ground truth, is captured by LSSInst. Besides, our methods yield a more consistent orientation and box scale with the ground truth in every view. In contrast, for example, there is a severe rotation shift (the red curved arrow) of the bus both in the $\mathtt{CAM\_FRONT}$ and $\mathtt{CAM\_FRONT\_LEFT}$ views as well as the box misalignment among those cars that are turning past the left traffic lights in the $\mathtt{CAM\_FRONT\_RIGHT}$ view. These observations above fully demonstrate the improvement of missing details, no matter the wider-range perception breadth or own more refined property.

\end{document}